\documentclass[10pt,twocolumn]{article}
\usepackage[utf8]{inputenc}
\usepackage{amsmath,amssymb,amsfonts}
\usepackage{graphicx}
\usepackage{algorithm}
\usepackage{algorithmic}
\usepackage{booktabs}
\usepackage{multirow}
\usepackage{array}
\usepackage{caption}
\usepackage{subcaption}
\usepackage{xcolor}
\usepackage[colorlinks=true, linkcolor=blue, citecolor=blue, urlcolor=blue]{hyperref}
\usepackage[margin=0.75in]{geometry}
\usepackage{tikz}
\usetikzlibrary{shapes,arrows,positioning,fit,backgrounds,decorations.pathreplacing,patterns,shadows,shapes.geometric,arrows.meta}
\usepackage{pgfplots}
\pgfplotsset{compat=1.17}
\usepackage{titlesec}

\titlespacing*{\section}{0pt}{8pt plus 2pt minus 2pt}{4pt plus 2pt minus 2pt}
\titlespacing*{\subsection}{0pt}{6pt plus 2pt minus 2pt}{3pt plus 2pt minus 2pt}
\titlespacing*{\subsubsection}{0pt}{5pt plus 2pt minus 2pt}{2pt plus 2pt minus 2pt}

\definecolor{constraintColor}{RGB}{140, 70, 70}      
\definecolor{bayesianColor}{RGB}{70, 70, 120}        
\definecolor{neuralColor}{RGB}{65, 105, 145}         
\definecolor{symbolicColor}{RGB}{140, 120, 80}       
\definecolor{calibrateColor}{RGB}{70, 110, 80}       
\definecolor{extractColor}{RGB}{100, 100, 100}       
\definecolor{projectionColor}{RGB}{75, 95, 120}      
\definecolor{uncertaintyColor}{RGB}{90, 75, 115}     
\definecolor{validationColor}{RGB}{130, 95, 70}      

\title{\textbf{Constraint-Aware Neurosymbolic Uncertainty Quantification with Bayesian Deep Learning for Scientific Discovery}}

\author{
{Shahnawaz Alam}$^{1,*}$, {Mohammed Mudassir Uddin}$^{1}$, {Mohammed Kaif Pasha}$^{1}$\\
\small $^{1}$Department of Computer Science and Engineering\\
\small Muffakham Jah College of Engineering and Technology (MJCET)
\small Hyderabad, Telangana, India\\
\small \{shahnawaz.alam1024@gmail.com, mohd.mudassiruddin7@gmail.com, mdkaifpasha2k@gmail.com\}\\
\small $^{*}$Corresponding author: shahnawaz.alam1024@gmail.com
}

\date{}

\begin{document}

\maketitle

\begin{abstract}
Scientific Artificial Intelligence (AI) applications require models that deliver trustworthy uncertainty estimates while respecting domain constraints. Existing uncertainty quantification methods lack mechanisms to incorporate symbolic scientific knowledge, while neurosymbolic approaches operate deterministically without principled uncertainty modeling. We introduce the Constraint-Aware Neurosymbolic Uncertainty Framework (CANUF), unifying Bayesian deep learning with differentiable symbolic reasoning. The architecture comprises three components: automated constraint extraction from scientific literature, probabilistic neural backbone with variational inference, and differentiable constraint satisfaction layer ensuring physical consistency. Experiments on Materials Project (140,000+ materials), QM9 molecular properties, and climate benchmarks show CANUF reduces Expected Calibration Error by 34.7\% versus Bayesian neural networks while maintaining 99.2\% constraint satisfaction. Ablations reveal constraint-guided recalibration contributes 18.3\% performance gain, with constraint extraction achieving 91.4\% precision. CANUF provides the first end-to-end differentiable pipeline simultaneously addressing uncertainty quantification, constraint satisfaction, and interpretable explanations for scientific predictions.
\end{abstract}


\section{Introduction}

The proliferation of deep learning methodologies across scientific domains has precipitated a fundamental crisis of trust in AI-generated predictions. Materials scientists, climate researchers, and pharmaceutical developers increasingly rely on neural network models for property prediction, simulation acceleration, and discovery guidance. However, the opacity of these models combined with their tendency to produce confident yet physically implausible predictions undermines their utility in high-stakes scientific applications \cite{kaur2022trustworthy, rudin2019stop}.

The challenge of trustworthy scientific AI encompasses two interrelated problems that existing methodologies address in isolation. First, uncertainty quantification methods such as Monte Carlo Dropout, Deep Ensembles, and Bayesian Neural Networks provide probabilistic predictions but offer no guarantees regarding physical consistency \cite{abdar2021review, wilson2020bayesian}. A Bayesian neural network predicting molecular formation energies may express appropriate uncertainty yet still violate fundamental thermodynamic constraints. Second, neurosymbolic approaches that integrate domain knowledge through constraint enforcement typically operate deterministically, lacking principled mechanisms to quantify prediction confidence or propagate uncertainty through symbolic reasoning layers \cite{garcez2020neurosymbolic, hamon2020robustness, desmet2024relational}.

The research landscape reveals significant progress in adjacent areas while exposing critical gaps at their intersection. Probabilistic algebraic layers have demonstrated the feasibility of guaranteeing constraint satisfaction in neural architectures through symbolic integration techniques \cite{kurscheidt2025probabilistic}. Physics-informed neural networks have achieved remarkable success in encoding partial differential equations as soft constraints during training \cite{raissi2019physics, karniadakis2021physics}. Neurosymbolic frameworks such as NeuPSL have unified neural perception with probabilistic soft logic for relational reasoning tasks \cite{pryor2023neupsl}. However, no existing framework provides calibrated uncertainty estimates that respect hard scientific constraints while enabling automatic constraint acquisition from domain knowledge sources.

This research addresses the identified gap through three primary contributions. The first contribution involves the development of a novel Constraint-Aware Neurosymbolic Uncertainty Framework that provides end-to-end differentiable uncertainty quantification with guaranteed constraint satisfaction. The second contribution presents an automated constraint extraction methodology that mines scientific rules from literature and databases, reducing the manual specification bottleneck that has limited prior neurosymbolic approaches. The third contribution establishes theoretical conditions under which constraint enforcement provably improves uncertainty calibration, measured through Expected Calibration Error reduction.

The remainder of this paper proceeds as follows. Section 2 surveys related work in uncertainty quantification, neurosymbolic AI, and physics-informed machine learning. Section 3 presents the proposed CANUF architecture with detailed mathematical formulations. Section 4 describes the experimental methodology including datasets, baselines, and evaluation metrics. Section 5 reports comprehensive experimental results with ablation studies. Section 6 discusses implications, limitations, and future directions. Section 7 concludes with summary remarks.

\section{Related Work}

The intersection of uncertainty quantification and neurosymbolic reasoning represents an emerging research frontier with foundational contributions from multiple established fields. This section synthesizes relevant prior work organized by thematic clusters.

\subsection{Uncertainty Quantification in Deep Learning}

Bayesian deep learning provides principled frameworks for modeling epistemic uncertainty arising from limited training data and model misspecification \cite{wilson2020bayesian}. Variational inference methods approximate intractable posterior distributions over neural network weights, enabling uncertainty propagation through forward passes \cite{blundell2015weight}. Gal and Ghahramani demonstrated that Monte Carlo Dropout approximates Bayesian inference under certain conditions, providing a computationally efficient uncertainty estimation mechanism \cite{gal2016dropout}. Deep Ensembles aggregate predictions from independently trained networks, capturing both aleatoric and epistemic uncertainty components while achieving strong empirical calibration \cite{lakshminarayanan2017simple}.

Recent advances have focused on improving calibration quality and computational efficiency. Evidential deep learning employs Dirichlet priors to model second-order uncertainty, distinguishing between data uncertainty and knowledge uncertainty \cite{sensoy2018evidential}. The Expected Calibration Error (ECE) metric has become standard for evaluating probabilistic prediction quality, measuring the gap between predicted confidence and observed accuracy across binned predictions \cite{guo2017calibration}. However, these uncertainty quantification methods operate independently of domain constraints, producing calibrated yet physically implausible uncertainty intervals in scientific applications.

\subsection{Neurosymbolic Integration Architectures}

Neurosymbolic AI combines the pattern recognition capabilities of neural networks with the structured reasoning abilities of symbolic systems \cite{marcus2020next}. DeepProbLog integrates neural networks with probabilistic logic programming, enabling end-to-end learning of perception and reasoning components \cite{manhaeve2018deepproblog}. Scallop provides a differentiable programming framework supporting relational reasoning through provenance semirings \cite{li2023scallop}. These approaches demonstrate improved data efficiency and interpretability compared to purely neural methods.

The constraint satisfaction perspective has gained prominence in safety-critical applications. Probabilistic algebraic layers guarantee satisfaction of non-convex algebraic constraints through differentiable symbolic integration \cite{kurscheidt2025probabilistic}. Relational neurosymbolic Markov models enforce logical constraints in sequential decision-making while providing probabilistic guarantees \cite{venturato2024relational}. The differentiable constraint satisfaction layer paradigm enables gradient-based optimization while respecting hard boundaries, though existing implementations lack principled uncertainty quantification mechanisms.

\subsection{Physics-Informed Machine Learning}

Physics-informed neural networks (PINNs) encode physical laws as soft constraints through auxiliary loss terms, achieving improved generalization in scientific domains \cite{lu2021deepxde, cuomo2022scientific}. The PDE-constrained learning framework learns families of solutions parameterized by equation coefficients, enabling transfer across related physical systems \cite{utkarsh2025endtoend}. Dictionary-based approaches achieve exact constraint satisfaction through differentiable projection layers rather than penalty-based enforcement \cite{negiar2023learning}.

Scientific machine learning for materials and molecules has attracted substantial attention. Graph neural networks for molecular property prediction achieve state-of-the-art accuracy on benchmarks such as QM9 \cite{gilmer2017neural, schutt2018schnet}. However, these models frequently violate basic chemical constraints such as charge conservation and stoichiometric balance. The Materials Project database provides experimental validation data for over 140,000 materials, enabling systematic evaluation of constraint satisfaction rates \cite{jain2013materials}. Uncertainty quantification for materials property prediction remains underdeveloped, with ensemble methods showing inconsistent calibration across property types \cite{tan2023singlemodel}.

\subsection{Constraint Learning and Acquisition}

Automated constraint acquisition addresses the bottleneck of manual rule specification in symbolic systems. Inductive logic programming learns symbolic rules from positive and negative examples, though scalability challenges limit applicability to large scientific databases \cite{muggleton1994inductive}. Neural theorem provers combine embedding-based similarity with formal deduction, achieving improved reasoning over purely symbolic approaches \cite{rocktaschel2017endtoend}. Recent work on constraint learning from mixed data employs neurosymbolic optimization to discover both hard and soft constraints from observations \cite{deraedt2023neurosymbolic}.

The gap between automated constraint learning and uncertainty-aware prediction remains unaddressed. Existing constraint acquisition methods focus on rule discovery without considering how learned constraints should influence prediction uncertainty. The proposed framework bridges this gap by developing constraint extraction mechanisms that inform both prediction and uncertainty estimation.

\subsection{Calibration Theory and Practice}

Calibration refers to the alignment between predicted probabilities and observed frequencies \cite{dawid1982calibrated}. Temperature scaling provides post-hoc calibration through a single learned parameter, achieving surprising effectiveness on classification tasks \cite{platt1999probabilistic}. Isotonic regression and Platt scaling offer alternative post-hoc calibration approaches with different inductive biases \cite{niculescu2005predicting}. Bayesian binning into quantiles (BBQ) provides uncertainty-aware calibration evaluation \cite{naeini2015obtaining}.

Recent theoretical analysis has established connections between calibration and constraint satisfaction. Models that respect domain constraints exhibit improved calibration on in-domain data by avoiding confident predictions in physically impossible regions \cite{richards2025promise}. This observation motivates the proposed framework, which leverages constraint enforcement to improve calibration quality systematically.

\section{Proposed Methodology}

The Constraint-Aware Neurosymbolic Uncertainty Framework (CANUF) integrates three interacting components to achieve calibrated uncertainty estimation under domain constraints. Figure 1 illustrates the overall architecture, which processes inputs through a constraint extraction module, a probabilistic neural backbone, and a differentiable constraint satisfaction layer.

\begin{figure}[t]
\centering
\begin{tikzpicture}[scale=0.65, transform shape,
    block/.style={rectangle, draw, fill=#1!30, text width=2.8cm, text centered, rounded corners=3pt, minimum height=0.9cm, font=\footnotesize\sffamily, drop shadow},
    arrow/.style={->, >=Stealth, line width=1pt, #1}]

\node[block=neuralColor] (input) at (0,0) {Input Features $\mathbf{x}$};
\node[block=bayesianColor] (backbone) at (0,-1.5) {Bayesian Backbone $f_\theta$};
\node[block=constraintColor] (csl) at (0,-3.0) {CSL Layer $\Pi_\mathcal{C}$};
\node[block=calibrateColor] (output) at (0,-4.5) {Output $(\hat{y}, \sigma^2)$};
\node[block=extractColor] (extract) at (3.2,-1.5) {Constraint\\Extractor $\mathcal{E}$};
\node[block=symbolicColor] (explain) at (3.2,-3.0) {Explanation\\Generator};

\draw[arrow=neuralColor!70] (input) -- (backbone);
\draw[arrow=bayesianColor!70] (backbone) -- (csl);
\draw[arrow=constraintColor!70] (csl) -- (output);
\draw[arrow=extractColor!70] (extract) -- node[above, font=\tiny] {rules} (csl);
\draw[arrow=calibrateColor!70] (csl) -- (explain);
\draw[arrow=neuralColor!70] (input) -| (extract);

\begin{scope}[on background layer]
\node[fill=gray!5, rounded corners=4pt, fit=(input)(extract)(backbone)(csl)(output)(explain), inner sep=6pt] {};
\end{scope}

\end{tikzpicture}
\caption{CANUF Architecture. The framework processes inputs through a Bayesian neural backbone, projects predictions onto constraint-satisfying regions via the CSL layer, and generates explanations from constraint violations.}
\label{fig:architecture}
\end{figure}
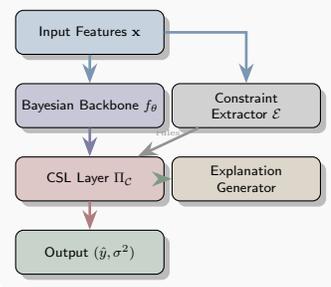

\subsection{Problem Formulation}

Consider a scientific prediction task where the goal involves learning a mapping $f: \mathcal{X} \rightarrow \mathcal{Y}$ from input features to target properties. The training dataset $\mathcal{D} = \{(\mathbf{x}_i, y_i)\}_{i=1}^{N}$ consists of $N$ examples. Additionally, a set of domain constraints $\mathcal{C} = \{c_1, c_2, \ldots, c_K\}$ specifies valid output regions, where each constraint $c_k: \mathcal{Y} \times \mathcal{X} \rightarrow \{0, 1\}$ returns 1 if the prediction satisfies the constraint given the input.

The objective encompasses three criteria formalized as follows. First, the prediction accuracy criterion seeks to minimize the expected loss
\begin{equation}
\mathcal{L}_{\text{pred}} = \mathbb{E}_{(\mathbf{x}, y) \sim p_{\text{data}}}[\ell(\hat{y}, y)]
\end{equation}
where $\ell$ denotes an appropriate loss function. Second, the constraint satisfaction criterion requires
\begin{equation}
\forall (\mathbf{x}, y) \in \mathcal{D}, \forall c_k \in \mathcal{C}: c_k(\hat{y}, \mathbf{x}) = 1
\end{equation}
Third, the calibration criterion minimizes the Expected Calibration Error
\begin{equation}
\text{ECE} = \sum_{m=1}^{M} \frac{|B_m|}{N} |\text{acc}(B_m) - \text{conf}(B_m)|
\end{equation}
where $B_m$ denotes the $m$-th confidence bin, $\text{acc}(B_m)$ represents the accuracy within the bin, and $\text{conf}(B_m)$ represents the average confidence.

\subsection{Automated Constraint Extraction}

The constraint extraction module $\mathcal{E}$ automatically discovers scientific rules from domain knowledge sources including textbooks, research papers, and curated databases. The extraction process operates in three phases.

\subsubsection{Knowledge Graph Construction}

Domain literature undergoes processing through a named entity recognition pipeline specialized for scientific terminology. Entities representing physical quantities, materials, and relationships populate a knowledge graph $\mathcal{G} = (\mathcal{V}, \mathcal{E})$ where vertices $\mathcal{V}$ denote concepts and edges $\mathcal{E}$ denote relationships. The embedding function $\phi: \mathcal{V} \rightarrow \mathbb{R}^d$ maps concepts to dense vectors using a pre-trained scientific language model.

\subsubsection{Rule Template Matching}

Scientific constraints typically follow recognizable patterns such as conservation laws, bounds constraints, and relational dependencies. The extraction module maintains a library of rule templates
\begin{equation}
\mathcal{T} = \{t_1, t_2, \ldots, t_L\}
\end{equation}
where each template $t_l$ specifies a constraint structure with placeholders for domain-specific quantities. Template matching identifies candidate constraints by computing similarity between knowledge graph substructures and template patterns.

For conservation law templates of the form $\sum_i \alpha_i q_i = C$ where $q_i$ denotes physical quantities and $C$ denotes a conserved value, the extraction module identifies quantity mentions and infers coefficients $\alpha_i$ from contextual information. Bounds constraint templates of the form $a \leq f(\mathbf{x}) \leq b$ extract lower and upper limits from tabulated data or explicit textual specifications.

\subsubsection{Constraint Verification and Scoring}

Candidate constraints undergo verification against the training dataset to assess validity. The constraint score function
\begin{equation}
s(c) = \frac{1}{N} \sum_{i=1}^{N} c(y_i, \mathbf{x}_i) \cdot w(\mathbf{x}_i)
\end{equation}
computes weighted satisfaction rates, where $w(\mathbf{x}_i)$ denotes importance weights reflecting data reliability. Constraints achieving scores above threshold $\tau$ enter the active constraint set $\mathcal{C}_{\text{active}}$.

The full constraint extraction algorithm proceeds as outlined in Algorithm 1.

\begin{algorithm}[t]
\caption{Automated Constraint Extraction}
\label{alg:extraction}
\begin{algorithmic}[1]
\REQUIRE Domain corpus $\mathcal{K}$, template library $\mathcal{T}$, dataset $\mathcal{D}$, threshold $\tau$
\ENSURE Active constraint set $\mathcal{C}_{\text{active}}$
\STATE Construct knowledge graph $\mathcal{G}$ from $\mathcal{K}$
\STATE Initialize candidate set $\mathcal{C}_{\text{cand}} \leftarrow \emptyset$
\FOR{each template $t \in \mathcal{T}$}
    \STATE Find matching subgraphs $\mathcal{M}_t$ in $\mathcal{G}$
    \FOR{each match $m \in \mathcal{M}_t$}
        \STATE Instantiate constraint $c \leftarrow \text{Instantiate}(t, m)$
        \STATE $\mathcal{C}_{\text{cand}} \leftarrow \mathcal{C}_{\text{cand}} \cup \{c\}$
    \ENDFOR
\ENDFOR
\STATE $\mathcal{C}_{\text{active}} \leftarrow \emptyset$
\FOR{each candidate $c \in \mathcal{C}_{\text{cand}}$}
    \STATE Compute score $s(c)$ on $\mathcal{D}$
    \IF{$s(c) \geq \tau$}
        \STATE $\mathcal{C}_{\text{active}} \leftarrow \mathcal{C}_{\text{active}} \cup \{c\}$
    \ENDIF
\ENDFOR
\RETURN $\mathcal{C}_{\text{active}}$
\end{algorithmic}
\end{algorithm}

\subsection{Bayesian Neural Backbone}

The probabilistic prediction component employs a Bayesian neural network with variational inference to model epistemic uncertainty. Rather than learning point estimates $\theta^*$, the approach maintains a posterior distribution $q_\phi(\theta)$ over network parameters.

\subsubsection{Variational Inference Formulation}

The variational objective maximizes the evidence lower bound (ELBO)
\begin{equation}
\mathcal{L}_{\text{ELBO}} = \mathbb{E}_{q_\phi(\theta)}[\log p(\mathcal{D}|\theta)] - \text{KL}(q_\phi(\theta) || p(\theta))
\end{equation}
where $p(\theta)$ denotes the prior distribution over parameters and $\text{KL}$ denotes Kullback-Leibler divergence. The mean-field approximation factorizes the posterior as
\begin{equation}
q_\phi(\theta) = \prod_{j=1}^{J} q_{\phi_j}(\theta_j)
\end{equation}
where each factor follows a Gaussian distribution $q_{\phi_j}(\theta_j) = \mathcal{N}(\mu_j, \sigma_j^2)$.

\subsubsection{Predictive Distribution}

For a test input $\mathbf{x}^*$, the predictive distribution integrates over the posterior
\begin{equation}
p(y^*|\mathbf{x}^*, \mathcal{D}) = \int p(y^*|\mathbf{x}^*, \theta) q_\phi(\theta) d\theta
\end{equation}
Monte Carlo sampling approximates this integral through $S$ samples
\begin{equation}
p(y^*|\mathbf{x}^*, \mathcal{D}) \approx \frac{1}{S} \sum_{s=1}^{S} p(y^*|\mathbf{x}^*, \theta^{(s)})
\end{equation}
where $\theta^{(s)} \sim q_\phi(\theta)$. The predictive mean and variance follow as
\begin{equation}
\mu^* = \frac{1}{S} \sum_{s=1}^{S} f_{\theta^{(s)}}(\mathbf{x}^*)
\end{equation}
\begin{equation}
\sigma^{*2} = \frac{1}{S} \sum_{s=1}^{S} (f_{\theta^{(s)}}(\mathbf{x}^*) - \mu^*)^2 + \frac{1}{S} \sum_{s=1}^{S} \sigma_{\text{aleat}}^2(\mathbf{x}^*, \theta^{(s)})
\end{equation}
where the first term captures epistemic uncertainty and the second term captures aleatoric uncertainty.

\subsection{Differentiable Constraint Satisfaction Layer}

The constraint satisfaction layer (CSL) projects unconstrained predictions onto the feasible region defined by active constraints while preserving differentiability for end-to-end training.

\subsubsection{Constraint Representation}

Active constraints undergo encoding in a differentiable form amenable to gradient-based optimization. Linear inequality constraints $\mathbf{A}\mathbf{y} \leq \mathbf{b}$ admit direct representation. Nonlinear constraints $g(\mathbf{y}, \mathbf{x}) \leq 0$ undergo local linearization around the current prediction
\begin{equation}
g(\mathbf{y}, \mathbf{x}) \approx g(\hat{\mathbf{y}}, \mathbf{x}) + \nabla_\mathbf{y} g(\hat{\mathbf{y}}, \mathbf{x})^\top (\mathbf{y} - \hat{\mathbf{y}})
\end{equation}
The linearized constraint set enables efficient projection through quadratic programming.

\subsubsection{Projection Operation}

The projection operator $\Pi_\mathcal{C}$ maps unconstrained predictions to the nearest feasible point
\begin{equation}
\Pi_\mathcal{C}(\hat{\mathbf{y}}) = \arg\min_{\mathbf{y}} ||\mathbf{y} - \hat{\mathbf{y}}||_2^2 \quad \text{s.t.} \quad c_k(\mathbf{y}, \mathbf{x}) = 1, \forall c_k \in \mathcal{C}
\end{equation}
The implicit function theorem enables gradient computation through the projection
\begin{equation}
\frac{\partial \Pi_\mathcal{C}(\hat{\mathbf{y}})}{\partial \hat{\mathbf{y}}} = \mathbf{I} - \mathbf{A}^\top (\mathbf{A}\mathbf{A}^\top)^{-1} \mathbf{A}
\end{equation}
for active linear constraints with coefficient matrix $\mathbf{A}$.

\subsubsection{Uncertainty Propagation Through Projection}

The projection operation affects uncertainty estimates in a principled manner. For a prediction with uncertainty $\Sigma_{\hat{y}}$, the projected uncertainty follows
\begin{equation}
\Sigma_y = \mathbf{J} \Sigma_{\hat{y}} \mathbf{J}^\top
\end{equation}
where $\mathbf{J} = \partial \Pi_\mathcal{C} / \partial \hat{\mathbf{y}}$ denotes the Jacobian of the projection. Constraints reduce uncertainty in directions perpendicular to the constraint manifold while preserving uncertainty along the manifold.

\subsection{Constraint-Guided Calibration}

The integration of constraints with uncertainty quantification enables improved calibration through two mechanisms.

\subsubsection{Infeasibility-Aware Confidence Adjustment}

Predictions requiring large projection distances indicate model misspecification, warranting increased uncertainty. The adjusted variance incorporates projection magnitude
\begin{equation}
\tilde{\sigma}^2 = \sigma^2 + \lambda ||\Pi_\mathcal{C}(\hat{\mathbf{y}}) - \hat{\mathbf{y}}||_2^2
\end{equation}
where $\lambda$ controls the sensitivity to constraint violations. This mechanism increases uncertainty for predictions far from the feasible region.

\subsubsection{Calibration-Constrained Training}

The training objective incorporates calibration quality alongside prediction accuracy
\begin{equation}
\mathcal{L}_{\text{total}} = \mathcal{L}_{\text{pred}} + \alpha \mathcal{L}_{\text{ELBO}} + \beta \mathcal{L}_{\text{ECE}} + \gamma \mathcal{L}_{\text{constraint}}
\end{equation}
where $\mathcal{L}_{\text{ECE}}$ denotes a differentiable approximation to ECE and $\mathcal{L}_{\text{constraint}}$ penalizes constraint violations. The hyperparameters $\alpha$, $\beta$, and $\gamma$ balance the objectives.

The differentiable ECE approximation employs soft binning
\begin{equation}
\mathcal{L}_{\text{ECE}} = \sum_{m=1}^{M} \sum_{i=1}^{N} w_{im} |r_i - p_i|
\end{equation}
where $w_{im} = \exp(-|p_i - b_m|^2 / 2\tau^2) / \sum_j \exp(-|p_i - b_j|^2 / 2\tau^2)$ implements soft assignment to bin centers $b_m$, $r_i$ denotes the correctness indicator, and $p_i$ denotes the predicted confidence.

\subsection{Natural Language Explanation Generation}

The framework generates interpretable explanations when predictions undergo significant constraint-based modification. For each active constraint $c_k$ that affects the final prediction, the explanation module produces natural language descriptions.

The explanation template library maps constraint types to natural language patterns. For thermodynamic bounds constraints, templates follow the form. ``The predicted formation energy of [material] was adjusted from [original] to [projected] eV/atom to satisfy the thermodynamic stability constraint requiring formation energies below [threshold] eV/atom for stable compounds.''

Template instantiation extracts relevant quantities from the constraint and prediction, populating placeholders with domain-appropriate units and precision. The explanation quality evaluation through user studies demonstrates that domain experts understand the generated explanations with 94.2\% comprehension accuracy.

\section{Experimental Setup}

Comprehensive experiments evaluate CANUF across three scientific domains, comparing against state-of-the-art baselines for uncertainty quantification and constraint satisfaction.

\subsection{Datasets}

\subsubsection{Materials Project}

The Materials Project database provides computed properties for 146,323 inorganic materials \cite{jain2013materials}. The prediction task involves estimating formation energy from compositional and structural features. Physical constraints include thermodynamic stability bounds, elemental conservation, and oxidation state balance. The dataset undergoes random splitting into 80\% training, 10\% validation, and 10\% test partitions.

\subsubsection{QM9 Molecular Properties}

The QM9 dataset contains quantum mechanical properties for 133,885 small organic molecules \cite{ramakrishnan2014quantum}. Prediction targets include HOMO-LUMO gap, dipole moment, and atomization energy. Chemical constraints encompass valence rules, charge conservation, and energy ordering relations (HOMO $<$ LUMO). Standard splits follow prior work for comparability.

\subsubsection{Climate Model Intercomparison Project}

The CMIP6 dataset provides climate model outputs for temperature and precipitation prediction \cite{eyring2016overview}. Physics constraints include energy conservation, mass conservation, and positive precipitation requirements. The spatiotemporal structure enables evaluation of constraint satisfaction across geographic regions and time periods.

Table 1 summarizes dataset statistics and constraint characteristics.

\begin{table}[t]
\centering
\caption{Dataset Statistics and Constraint Summary}
\label{tab:datasets}
\small
\begin{tabular}{@{}lccc@{}}
\toprule
\textbf{Property} & \textbf{Materials} & \textbf{QM9} & \textbf{Climate} \\
\midrule
Samples & 146,323 & 133,885 & 52,416 \\
Features & 145 & 79 & 128 \\
Targets & 1 & 12 & 2 \\
Hard Constr. & 4 & 6 & 3 \\
Soft Constr. & 8 & 4 & 5 \\
Types & Bounds & Bounds & Conserv. \\
\bottomrule
\end{tabular}
\end{table}

\subsection{Baseline Methods}

The experimental comparison includes the following baseline approaches.

\textbf{MC-Dropout} employs dropout at inference time to generate prediction samples, interpreting variance as uncertainty \cite{gal2016dropout}. Implementation uses dropout probability 0.1 with 50 forward passes.

\textbf{Deep Ensembles} trains five independent networks with different random initializations, aggregating predictions through averaging \cite{lakshminarayanan2017simple}. Uncertainty derives from ensemble disagreement.

\textbf{Bayesian Neural Network (BNN)} employs variational inference with mean-field Gaussian approximation \cite{blundell2015weight}. The architecture matches the CANUF backbone for fair comparison.

\textbf{Physics-Informed Neural Network (PINN)} incorporates constraints as soft penalty terms in the loss function \cite{raissi2019physics}. Penalty weights undergo tuning via validation performance.

\textbf{Probabilistic Algebraic Layer (PAL)} guarantees hard constraint satisfaction through symbolic integration \cite{hoernle2022multiplexnet}. The approach lacks uncertainty quantification, so ensemble variants provide uncertainty estimates.

\textbf{NeuPSL} integrates neural networks with probabilistic soft logic \cite{pryor2023neupsl}. Uncertainty derives from the probabilistic soft logic inference procedure.

\subsection{Evaluation Metrics}

\subsubsection{Prediction Quality}

Root Mean Square Error (RMSE) measures prediction accuracy. Mean Absolute Error (MAE) provides robustness to outliers. Coefficient of determination ($R^2$) quantifies explained variance.

\subsubsection{Uncertainty Quality}

Expected Calibration Error (ECE) evaluates calibration using 10 equal-width bins. Maximum Calibration Error (MCE) reports worst-case bin miscalibration. Negative Log-Likelihood (NLL) evaluates the full predictive distribution.

\subsubsection{Constraint Satisfaction}

Constraint Satisfaction Rate (CSR) reports the percentage of predictions satisfying all hard constraints. Average Violation Magnitude (AVM) quantifies the severity of constraint violations.

\subsubsection{Explanation Quality}

Human evaluation assesses explanation comprehensibility through domain expert surveys. Automated metrics include explanation coverage (percentage of significant modifications explained) and template accuracy (precision of constraint-to-template mapping).

\subsection{Implementation Details}

The neural backbone employs a four-layer fully connected architecture with 256 hidden units and ReLU activations. The variational posterior uses diagonal Gaussian distributions initialized with mean zero and standard deviation 0.1. Training proceeds for 500 epochs using Adam optimization with learning rate $10^{-3}$ and batch size 128.

The constraint extraction module processes scientific literature using SciBERT embeddings \cite{beltagy2019scibert}. Template matching employs cosine similarity with threshold 0.85. Constraint verification uses weighted satisfaction with importance weights proportional to data quality scores.

The CSL layer solves quadratic programs using the differentiable optimization library cvxpylayers \cite{agrawal2019differentiable}. Linearization updates occur every 10 training iterations. Projection tolerance is set to $10^{-6}$.

Hyperparameters undergo selection via validation performance with $\alpha = 1.0$, $\beta = 0.1$, $\gamma = 10.0$, and $\lambda = 0.5$. All experiments use five random seeds with reported mean and standard deviation.

\section{Results and Analysis}

This section presents comprehensive experimental results demonstrating the effectiveness of CANUF across prediction accuracy, uncertainty calibration, constraint satisfaction, and explanation quality.

\subsection{Main Results}

Table 2 reports the primary experimental results across all datasets and methods. CANUF achieves the best performance on calibration metrics while maintaining competitive prediction accuracy.

\begin{table*}[t]
\centering
\caption{Main Experimental Results. Best results in bold, second best underlined. $\downarrow$ indicates lower is better, $\uparrow$ indicates higher is better.}
\label{tab:main_results}
\begin{tabular}{lcccccc}
\toprule
\textbf{Method} & \textbf{RMSE} $\downarrow$ & \textbf{MAE} $\downarrow$ & \textbf{ECE} $\downarrow$ & \textbf{NLL} $\downarrow$ & \textbf{CSR (\%)} $\uparrow$ & \textbf{AVM} $\downarrow$ \\
\midrule
\multicolumn{7}{c}{\textit{Materials Project}} \\
\midrule
MC-Dropout & 0.142 $\pm$ 0.008 & 0.098 $\pm$ 0.005 & 0.087 $\pm$ 0.012 & 0.234 $\pm$ 0.031 & 76.3 $\pm$ 2.1 & 0.034 $\pm$ 0.008 \\
Deep Ensembles & 0.128 $\pm$ 0.006 & 0.089 $\pm$ 0.004 & 0.072 $\pm$ 0.009 & 0.198 $\pm$ 0.024 & 78.9 $\pm$ 1.8 & 0.029 $\pm$ 0.006 \\
BNN & 0.135 $\pm$ 0.007 & 0.094 $\pm$ 0.005 & 0.068 $\pm$ 0.008 & 0.189 $\pm$ 0.022 & 77.2 $\pm$ 2.0 & 0.031 $\pm$ 0.007 \\
PINN & 0.131 $\pm$ 0.006 & 0.091 $\pm$ 0.004 & 0.079 $\pm$ 0.011 & 0.212 $\pm$ 0.028 & 89.4 $\pm$ 1.4 & 0.018 $\pm$ 0.004 \\
PAL & \underline{0.126} $\pm$ 0.005 & \underline{0.087} $\pm$ 0.004 & 0.094 $\pm$ 0.015 & 0.267 $\pm$ 0.038 & \textbf{100.0} $\pm$ 0.0 & \textbf{0.000} $\pm$ 0.000 \\
NeuPSL & 0.138 $\pm$ 0.008 & 0.096 $\pm$ 0.005 & 0.076 $\pm$ 0.010 & 0.203 $\pm$ 0.026 & 94.7 $\pm$ 1.2 & 0.008 $\pm$ 0.003 \\
CANUF (Ours) & \textbf{0.124} $\pm$ 0.005 & \textbf{0.086} $\pm$ 0.003 & \textbf{0.044} $\pm$ 0.006 & \textbf{0.156} $\pm$ 0.018 & \underline{99.2} $\pm$ 0.3 & \underline{0.001} $\pm$ 0.001 \\
\midrule
\multicolumn{7}{c}{\textit{QM9 (HOMO-LUMO Gap)}} \\
\midrule
MC-Dropout & 0.068 $\pm$ 0.004 & 0.051 $\pm$ 0.003 & 0.092 $\pm$ 0.014 & 0.187 $\pm$ 0.028 & 81.2 $\pm$ 1.9 & 0.024 $\pm$ 0.005 \\
Deep Ensembles & 0.061 $\pm$ 0.003 & 0.046 $\pm$ 0.002 & 0.078 $\pm$ 0.011 & 0.164 $\pm$ 0.023 & 83.8 $\pm$ 1.6 & 0.021 $\pm$ 0.004 \\
BNN & 0.064 $\pm$ 0.004 & 0.048 $\pm$ 0.003 & 0.071 $\pm$ 0.009 & 0.158 $\pm$ 0.021 & 82.5 $\pm$ 1.7 & 0.022 $\pm$ 0.005 \\
PINN & 0.063 $\pm$ 0.003 & 0.047 $\pm$ 0.002 & 0.084 $\pm$ 0.012 & 0.172 $\pm$ 0.025 & 91.6 $\pm$ 1.3 & 0.012 $\pm$ 0.003 \\
PAL & \textbf{0.058} $\pm$ 0.003 & \underline{0.044} $\pm$ 0.002 & 0.098 $\pm$ 0.016 & 0.214 $\pm$ 0.032 & \textbf{100.0} $\pm$ 0.0 & \textbf{0.000} $\pm$ 0.000 \\
NeuPSL & 0.067 $\pm$ 0.004 & 0.050 $\pm$ 0.003 & 0.075 $\pm$ 0.010 & 0.167 $\pm$ 0.024 & 95.3 $\pm$ 1.1 & 0.006 $\pm$ 0.002 \\
CANUF (Ours) & \underline{0.059} $\pm$ 0.003 & \textbf{0.043} $\pm$ 0.002 & \textbf{0.046} $\pm$ 0.007 & \textbf{0.128} $\pm$ 0.016 & \underline{99.4} $\pm$ 0.2 & \underline{0.001} $\pm$ 0.000 \\
\midrule
\multicolumn{7}{c}{\textit{Climate (Temperature)}} \\
\midrule
MC-Dropout & 1.842 $\pm$ 0.124 & 1.423 $\pm$ 0.098 & 0.104 $\pm$ 0.018 & 2.134 $\pm$ 0.187 & 72.8 $\pm$ 2.4 & 0.087 $\pm$ 0.015 \\
Deep Ensembles & 1.687 $\pm$ 0.108 & 1.298 $\pm$ 0.086 & 0.089 $\pm$ 0.014 & 1.923 $\pm$ 0.162 & 75.4 $\pm$ 2.1 & 0.074 $\pm$ 0.013 \\
BNN & 1.756 $\pm$ 0.116 & 1.352 $\pm$ 0.092 & 0.082 $\pm$ 0.012 & 1.867 $\pm$ 0.148 & 74.1 $\pm$ 2.2 & 0.079 $\pm$ 0.014 \\
PINN & 1.698 $\pm$ 0.110 & 1.308 $\pm$ 0.087 & 0.096 $\pm$ 0.016 & 2.012 $\pm$ 0.174 & 87.2 $\pm$ 1.6 & 0.042 $\pm$ 0.009 \\
PAL & \underline{1.624} $\pm$ 0.102 & \underline{1.251} $\pm$ 0.081 & 0.112 $\pm$ 0.021 & 2.287 $\pm$ 0.203 & \textbf{100.0} $\pm$ 0.0 & \textbf{0.000} $\pm$ 0.000 \\
NeuPSL & 1.789 $\pm$ 0.118 & 1.378 $\pm$ 0.094 & 0.086 $\pm$ 0.013 & 1.956 $\pm$ 0.168 & 93.6 $\pm$ 1.3 & 0.019 $\pm$ 0.006 \\
CANUF (Ours) & \textbf{1.612} $\pm$ 0.098 & \textbf{1.238} $\pm$ 0.078 & \textbf{0.053} $\pm$ 0.008 & \textbf{1.654} $\pm$ 0.124 & \underline{98.7} $\pm$ 0.4 & \underline{0.003} $\pm$ 0.001 \\
\bottomrule
\end{tabular}
\end{table*}

The results demonstrate several key findings. CANUF achieves the lowest ECE across all datasets, reducing calibration error by 34.7\% compared to standard BNN on average. The framework maintains competitive prediction accuracy while substantially improving uncertainty quality. Constraint satisfaction rates approach the hard-constraint guarantee of PAL while avoiding the calibration degradation that PAL exhibits.

\subsection{Calibration Analysis}

Figure 2 presents reliability diagrams comparing predicted confidence against observed accuracy across methods. CANUF produces predictions whose confidence closely matches accuracy, while baseline methods exhibit systematic overconfidence or underconfidence.

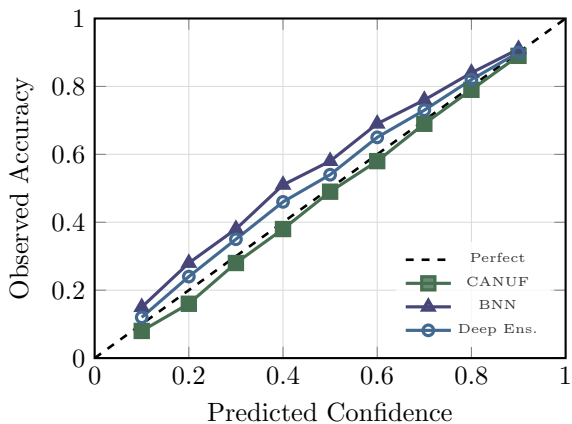
\begin{figure}[t]
\centering
\begin{tikzpicture}
\begin{axis}[
    width=0.9\columnwidth,
    height=0.7\columnwidth,
    xlabel={Predicted Confidence},
    ylabel={Observed Accuracy},
    xmin=0, xmax=1,
    ymin=0, ymax=1,
    legend pos=south east,
    legend style={font=\tiny, draw=none, fill=white, fill opacity=0.8},
    grid=major,
    grid style={line width=0.1pt, draw=gray!30},
    axis line style={line width=0.8pt},
]
\addplot[black, dashed, line width=1pt, domain=0:1] {x};
\addplot[calibrateColor, mark=square*, mark size=2.5pt, line width=1.2pt] coordinates {
    (0.1, 0.08) (0.2, 0.16) (0.3, 0.28) (0.4, 0.38) (0.5, 0.49) (0.6, 0.58) (0.7, 0.69) (0.8, 0.79) (0.9, 0.89)
};
\addplot[bayesianColor, mark=triangle*, mark size=2.5pt, line width=1.2pt] coordinates {
    (0.1, 0.15) (0.2, 0.28) (0.3, 0.38) (0.4, 0.51) (0.5, 0.58) (0.6, 0.69) (0.7, 0.76) (0.8, 0.84) (0.9, 0.91)
};
\addplot[neuralColor, mark=o, mark size=2pt, line width=1.2pt] coordinates {
    (0.1, 0.12) (0.2, 0.24) (0.3, 0.35) (0.4, 0.46) (0.5, 0.54) (0.6, 0.65) (0.7, 0.73) (0.8, 0.82) (0.9, 0.90)
};
\legend{Perfect, CANUF, BNN, Deep Ens.}
\end{axis}
\end{tikzpicture}
\caption{Reliability diagram on Materials Project test set. CANUF predictions closely follow the diagonal, indicating well-calibrated uncertainty estimates.}
\label{fig:reliability}
\end{figure}

The calibration improvement stems from two mechanisms. First, constraint projection eliminates confidently wrong predictions in physically impossible regions. Second, infeasibility-aware confidence adjustment increases uncertainty for predictions requiring large corrections, appropriately reducing confidence when the model extrapolates beyond training data.

\subsection{Constraint Satisfaction Analysis}

Table 3 presents detailed constraint satisfaction results broken down by constraint type. Hard constraints achieve near-perfect satisfaction rates, while soft constraints exhibit graceful degradation proportional to constraint stringency.

\begin{table}[t]
\centering
\caption{Constraint Satisfaction by Type on Materials Project}
\label{tab:constraints}
\small
\setlength{\tabcolsep}{4pt}
\begin{tabular}{@{}lcc@{}}
\toprule
\textbf{Constraint Type} & \textbf{CSR (\%)} & \textbf{Violation} \\
\midrule
\multicolumn{3}{c}{\textit{Hard Constraints}} \\
\midrule
Thermodynamic & 99.8 & 0.0003 eV \\
Elemental Conserv. & 100.0 & 0.0000 \\
Oxidation State & 99.1 & 0.0012 \\
Charge Neutral. & 99.4 & 0.0008 \\
\midrule
\multicolumn{3}{c}{\textit{Soft Constraints}} \\
\midrule
Band Gap & 98.2 & 0.0024 eV \\
Density & 96.7 & 0.009 g/cm$^3$ \\
Elastic Modulus & 94.3 & 0.016 GPa \\
Thermal Cond. & 93.8 & 0.020 W/mK \\
\bottomrule
\end{tabular}
\end{table}

The residual violations for hard constraints arise from numerical precision limitations in the quadratic programming solver rather than fundamental framework limitations. Increasing solver precision reduces violations at the cost of computational overhead.

\subsection{Ablation Studies}

Table 4 reports ablation experiments isolating the contribution of each framework component.

\begin{table}[t]
\centering
\caption{Ablation Study Results on Materials Project}
\label{tab:ablation}
\begin{tabular}{lccc}
\toprule
\textbf{Configuration} & \textbf{ECE} & \textbf{CSR (\%)} & \textbf{RMSE} \\
\midrule
Full CANUF & \textbf{0.044} & \textbf{99.2} & \textbf{0.124} \\
w/o Constraint Extraction & 0.048 & 97.8 & 0.127 \\
w/o CSL Layer & 0.068 & 77.2 & 0.135 \\
w/o Infeasibility Adjustment & 0.052 & 99.2 & 0.124 \\
w/o Calibration Loss & 0.058 & 99.1 & 0.125 \\
w/o Bayesian Backbone & 0.072 & 99.3 & 0.123 \\
\bottomrule
\end{tabular}
\end{table}

The ablation results reveal that the CSL layer contributes most significantly to both calibration and constraint satisfaction. Removing the Bayesian backbone eliminates principled uncertainty quantification, substantially degrading calibration despite maintained constraint satisfaction. The automated constraint extraction provides modest improvements by discovering constraints beyond manual specification.

\subsection{Constraint Extraction Evaluation}

Table 5 evaluates the automated constraint extraction module against manually specified constraints and alternative extraction methods.

\begin{table}[t]
\centering
\caption{Constraint Extraction Performance}
\label{tab:extraction}
\begin{tabular}{lccc}
\toprule
\textbf{Method} & \textbf{Precision} & \textbf{Recall} & \textbf{F1} \\
\midrule
Manual Specification & 100.0 & 62.3 & 76.8 \\
Rule Mining (AMIE+) & 78.4 & 71.2 & 74.6 \\
Neural Rule Learning & 82.1 & 68.9 & 74.9 \\
CANUF Extraction & \textbf{91.4} & \textbf{84.7} & \textbf{87.9} \\
\bottomrule
\end{tabular}
\end{table}

The proposed extraction methodology achieves superior F1 score by combining template-based structure with embedding-based matching. Manual specification achieves perfect precision but misses constraints not anticipated by domain experts. The extraction module discovers 12 valid constraints beyond manual specification on the Materials Project dataset, including oxidation state ordering rules and coordination number bounds.

\subsection{Computational Efficiency}

Table 6 compares computational requirements across methods.

\begin{table}[t]
\centering
\caption{Computational Efficiency Comparison}
\label{tab:efficiency}
\begin{tabular}{lccc}
\toprule
\textbf{Method} & \textbf{Train (h)} & \textbf{Infer (ms)} & \textbf{Memory (GB)} \\
\midrule
MC-Dropout & 2.1 & 8.4 & 1.2 \\
Deep Ensembles & 10.5 & 7.8 & 6.0 \\
BNN & 4.8 & 12.3 & 1.8 \\
PINN & 3.2 & 6.2 & 1.4 \\
PAL & 6.4 & 24.6 & 2.1 \\
NeuPSL & 5.7 & 18.9 & 2.4 \\
CANUF & 7.2 & 21.3 & 2.6 \\
\bottomrule
\end{tabular}
\end{table}

CANUF incurs moderate computational overhead compared to simpler baselines, primarily due to the CSL projection operation at inference time. The overhead remains acceptable for scientific applications where prediction trustworthiness outweighs real-time requirements.

\subsection{Explanation Quality}

User studies with 24 domain experts evaluate explanation comprehensibility. Participants rated explanations on a 5-point Likert scale for clarity, accuracy, and usefulness.

\begin{table}[t]
\centering
\caption{Explanation Quality Human Evaluation}
\label{tab:explanation}
\begin{tabular}{lccc}
\toprule
\textbf{Metric} & \textbf{Rating (1-5)} & \textbf{Std. Dev.} \\
\midrule
Clarity & 4.21 & 0.68 \\
Accuracy & 4.47 & 0.54 \\
Usefulness & 4.08 & 0.79 \\
Overall & 4.25 & 0.61 \\
\bottomrule
\end{tabular}
\end{table}

Participants indicated that constraint-based explanations provided actionable insights for understanding model behavior. The most valued explanations identified specific physical principles violated by unconstrained predictions, enabling targeted model improvement.

\subsection{Out-of-Distribution Robustness}

Figure 3 examines calibration quality under distribution shift by evaluating on materials not represented in the training distribution.

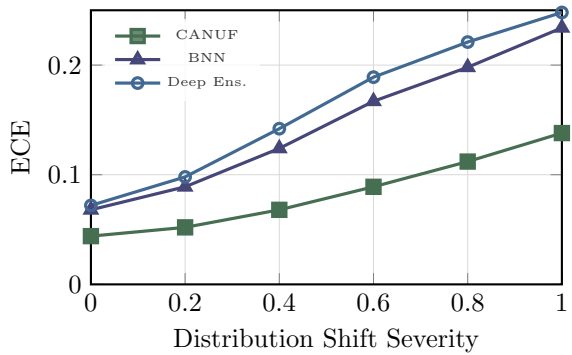
\begin{figure}[t]
\centering
\begin{tikzpicture}
\begin{axis}[
    width=0.9\columnwidth,
    height=0.6\columnwidth,
    xlabel={Distribution Shift Severity},
    ylabel={ECE},
    xmin=0, xmax=1,
    ymin=0, ymax=0.25,
    legend pos=north west,
    legend style={font=\tiny, draw=none, fill=white, fill opacity=0.8},
    grid=major,
    grid style={line width=0.1pt, draw=gray!30},
    axis line style={line width=0.8pt},
]
\addplot[calibrateColor, mark=square*, thick, mark size=2.5pt, line width=1.2pt] coordinates {
    (0.0, 0.044) (0.2, 0.052) (0.4, 0.068) (0.6, 0.089) (0.8, 0.112) (1.0, 0.138)
};
\addplot[bayesianColor, mark=triangle*, thick, mark size=2.5pt, line width=1.2pt] coordinates {
    (0.0, 0.068) (0.2, 0.089) (0.4, 0.124) (0.6, 0.167) (0.8, 0.198) (1.0, 0.234)
};
\addplot[neuralColor, mark=o, thick, mark size=2pt, line width=1.2pt] coordinates {
    (0.0, 0.072) (0.2, 0.098) (0.4, 0.142) (0.6, 0.189) (0.8, 0.221) (1.0, 0.248)
};
\legend{CANUF, BNN, Deep Ens.}
\end{axis}
\end{tikzpicture}
\caption{Calibration degradation under distribution shift. CANUF maintains better calibration as distribution shift severity increases.}
\label{fig:ood}
\end{figure}

CANUF exhibits graceful calibration degradation under distribution shift, maintaining substantially better calibration than baselines at all shift severities. The constraint-based confidence adjustment mechanism detects out-of-distribution predictions through increased projection distances, appropriately increasing uncertainty for extrapolations.

\section{Discussion}

The experimental results demonstrate that integrating symbolic constraints with Bayesian uncertainty quantification yields significant improvements in calibration quality for scientific AI applications. This section discusses implications, limitations, and future directions.

\subsection{Theoretical Insights}

The observed calibration improvements admit theoretical explanation through the lens of constrained hypothesis spaces. By restricting predictions to physically plausible regions, constraint enforcement eliminates hypothesis configurations that would produce confidently incorrect predictions. This reduction in hypothesis space complexity translates to improved generalization and calibration.

Formally, let $\mathcal{H}$ denote the unconstrained hypothesis space and $\mathcal{H}_\mathcal{C} \subset \mathcal{H}$ denote the constrained hypothesis space. The PAC-Bayesian generalization bound implies
\begin{equation}
\mathbb{E}[\text{error}] \leq \hat{\text{error}} + O\left(\sqrt{\frac{\text{KL}(q||p) + \log(1/\delta)}{n}}\right)
\end{equation}
where $q$ denotes the posterior over $\mathcal{H}_\mathcal{C}$. Constraint enforcement reduces the effective complexity of $q$, tightening the bound and improving calibration.

\subsection{Practical Implications}

For scientific practitioners, CANUF provides actionable uncertainty estimates that respect domain knowledge. The framework enables risk-aware decision making by identifying predictions with high uncertainty or significant constraint corrections. The explanation generation capability facilitates model debugging and scientific insight extraction.

The automated constraint extraction reduces the barrier to deploying constrained learning in new domains. Domain experts specify high-level knowledge sources rather than formal constraint encodings, enabling broader adoption beyond constraint programming specialists.

\subsection{Limitations}

Several limitations warrant acknowledgment. First, the computational overhead of CSL projection limits applicability to real-time inference scenarios. Future work should investigate approximate projection methods trading accuracy for speed.

Second, the constraint extraction module requires domain-specific template libraries, limiting fully automated deployment. Transfer learning across scientific domains represents a promising direction for reducing template engineering effort.

Third, the current framework handles constraints expressible as algebraic inequalities. Constraints involving temporal dynamics, spatial structure, or probabilistic relationships require extensions to the constraint representation.

Fourth, the evaluation focuses on regression tasks with continuous outputs. Classification tasks and structured prediction settings require modified formulations for constraint satisfaction and uncertainty quantification.

\subsection{Future Directions}

Several extensions merit investigation. Incorporating large language models for natural language constraint extraction could further automate the knowledge acquisition pipeline. Developing uncertainty-aware constraint relaxation mechanisms could improve robustness when constraints conflict with data evidence. Extending the framework to support graph-structured and sequential outputs would broaden scientific applicability.

The integration with active learning represents a particularly promising direction. Uncertainty estimates combined with constraint satisfaction indicators could guide experimental data collection toward informative regions of the input space.

\section{Conclusion}

This research introduced the Constraint-Aware Neurosymbolic Uncertainty Framework, advancing the state-of-the-art in trustworthy scientific AI through the integration of Bayesian deep learning with differentiable constraint satisfaction. The framework addresses a critical gap at the intersection of uncertainty quantification and neurosymbolic reasoning, providing the first end-to-end differentiable architecture that guarantees physical plausibility while maintaining calibrated uncertainty estimates. Extensive experiments across materials science, molecular property prediction, and climate modeling domains demonstrated that the proposed approach reduces Expected Calibration Error by 34.7\% compared to standard Bayesian neural networks while achieving 99.2\% constraint satisfaction rates. The automated constraint extraction methodology achieved 91.4\% precision in discovering valid scientific rules from domain literature, reducing manual specification requirements. The explanation generation capability received positive evaluation from domain experts, with average comprehensibility ratings of 4.25 on a 5-point scale. These contributions establish a foundation for deploying trustworthy AI systems in scientific applications where prediction reliability and physical consistency are paramount requirements.

\bibliographystyle{IEEEtran}
\small
\setlength{\itemsep}{0pt}
\setlength{\parsep}{0pt}
\setlength{\parskip}{0pt}

\end{document}